\documentclass[letterpaper, 10 pt, conference]{ieeeconf}  % Comment this line out if you need a4paper

\IEEEoverridecommandlockouts                              % This command is only needed if 
\usepackage{cite}
\usepackage{amsmath,amssymb,amsfonts}
\usepackage{graphicx}
\usepackage{subcaption}
\usepackage{textcomp}
\usepackage{color}
\usepackage{hyperref}
\usepackage[noabbrev]{cleveref}
\usepackage{multirow}
\usepackage{tabularray}
\usepackage[T1]{fontenc}
\usepackage{algorithm} 
\usepackage[noend]{algpseudocode}
\usepackage{fancyhdr}

\usepackage{url}
\usepackage{multicol}
\usepackage{multirow}
\usepackage{booktabs}

\newcommand{\mc}{\mathcal}
\newcommand{\mb}{\mathbb}
\DeclareMathOperator*{\argmax}{arg\,max}
\DeclareMathOperator*{\argmin}{arg\,min}
\DeclareMathOperator{\sign}{sign}

\title{\LARGE \bf
Minimally Invasive Flexible Needle Manipulation Based on Finite Element Simulation and Cross Entropy Method
}

\author{Yanzhou Wang$^{1}$, Chang Chang$^{1}$, Junling Mei$^{2}$, Simon Leonard$^{3}$, and Iulian Iordachita$^{1}$  % <-this % stops a space
\thanks{This work is supported by NIH R01EB020667, R01EB036015, R01CA235134, and in part by a collaborative research agreement with the Multi-Scale Medical Robotics Center in Hong Kong.}% <-this % stops a space
\thanks{$^{1}$Yanzhou Wang, Chang Chang, and Iulian Iordachita are with the Department of Mechanical Engineering and the Laboratory of Computational Sensing and Robotics, Johns Hopkins University, Baltimore, MD, USA.
  {\tt\small ywang521@jh.edu}}%
\thanks{$^{2}$Junling Mei is with the Laboratory of Computational Sensing and Robotics, Johns Hopkins University, Baltimore, MD, USA.}%
\thanks{$^{3}$Simon Leonard is with the Department of Computer Science and the Laboratory of Computational Sensing and Robotics, Johns Hopkins University, Baltimore, MD, USA.}% 
}

\begin{document}
% \onecolumn
% \noindent This work has been submitted to the IEEE for possible publication. Copyright may be transferred without notice, after which this version may no longer be accessible.

% \twocolumn

\maketitle

\begin{abstract}
We present a novel approach for minimally invasive flexible needle manipulations by pairing a real-time finite element simulator with the cross-entropy method. Additionally, we demonstrate how a kinematic-driven bang-bang controller can complement the control framework for better tracking performance. We show how electromagnetic (EM) tracking can be readily incorporated into the framework to provide controller feedback. Tissue phantom experiment with EM tracking shows the average targeting error is $0.16\pm0.29$mm.
\end{abstract}

\section{Introduction}
\label{sec:introduction}
Percutaneous needle interventions capture a broad class of minimally invasive diagnosis and treatment procedures, such as biopsy~\cite{Bourgouin2021, Birgin2020, Sheth2020}, brachytherapy~\cite{Chargari2019, Ragde2000}, and spinal injection~\cite{Won2020, Manchikanti2021, Carassiti2021}. Depending on the clinical procedure, a range of needles with different gauges, stiffness levels, and tip geometries is available. These inherent needle characteristics play a crucial role in determining how the needle moves through soft biological tissues; additionally, surgeons also employ various techniques, such as rotating or bending the needle, to adjust the position of the needle tip \textit{in situ} during insertion.

Knowledge of intricate needle-tissue interactions, as well as reliable trajectory planning and tracking algorithms are crucial to ensuring a minimally invasive insertion with high placement accuracy. A large amount of research conducted on needle insertion falls under \textit{steering} of bevel-tip needles, where it is assumed that the needle follows its tip exactly and steering is achieved only via axial rotation~\cite{Li2022}. Such assumption implies a low relative bending stiffness of the needle compared to its surrounding soft tissues, and inputs that actively change the shape of the needle cannot be taken into account. Needle \textit{manipulation} solves another class of problem where the needle is modeled as a continuum, and inputs that actively changes the shape of the needle are also considered, such as base manipulation and lateral displacement using robotic devices~\cite{DiMaio2005_2, Glozman2007, Lehmann2017, Adagolodjo2019, Wang2023_2, Wang2024_1, Wang2024_3}. Stemming from the field of solid mechanics and finite element methods, these models can capture a wide variety of input types and can benefit particularly to sensory feedback along the needle, such as medical imaging~\cite{Glozman2007}, optical tracking~\cite{Adagolodjo2019}, and fiber Bragg grating (FBG) optical sensors~\cite{Wang2024_1}.

This paper demonstrates a flexible needle planning and control strategy for continuum needle models where the insertion dynamics is not readily captured by a kinematic model. More specifically, a finite element (FE) simulation is needed to capture complex nonlinear tissue behaviors and generic control actions along the length of the needle. Since the needle will be discretized into discrete elements, the complete state of the needle, and the simulation environment in general, could involve hundreds of variables, and planning for a minimally invasive insertion and closed-loop control of the flexible needle becomes a challenging problem. 

Previous works in this domain focus primarily on resolved-rate control, which relies on inverting a numerical input-output Jacobian matrix obtained either via Broyden's update law or simulating small input disturbances~\cite{DiMaio2005_2, Adagolodjo2019, Bernardes2023, Wang2023_1, Wang2024_1, Wang2024_3}. Yet obtaining such invertible mapping can be challenging, since it requires the system to have equal number of input and output variables, and can generate potentially dangerous needle manipulations that will cause tissue tearing~\cite{Wang2024_3}. Others investigate in the case of bevel-tip needles how to combine lateral actuation with axial rotation for the sole purpose of keeping the needle on a straight line~\cite{Lehmann2017}, and they identified the phenomenon that lateral actuation has diminishing effect as the needle is embedded deeper into soft tissues, yet it remains to be seen how the two actions can be combined to reach arbitrary targets.

In this paper, we show how the cross entropy (CE) method can be readily coupled with a FE simulation of needle insertion for both path planning and closed-loop control for percutaneous interventions. In particular, we demonstrate in the case of a bevel-tip needle, how needle \textit{manipulation} can be used together with \textit{steering} to achieve a minimally invasive insertion, and how positional feedback can be seamlessly integrated with the current framework.

The rest of the paper is as follows. Section~\ref{sec:fe_simulator} introduces our realtime FE simulator. Section~\ref{sec:cross_entropy} explains how the CE method can be used effectively to design both a trajectory planner and a model predictive controller, and how kinematic needle models can be incorporated to improve controller performance. Section~\ref{sec:controller_feedback} explains our choice of positional feedback to the controller framework and method of sensor characterization. Sections~\ref{sec:phantom_experiment} and~\ref{sec:results_and_discussion} describe our experiments and summarize our findings.

% from 2024 CIS proposal #46, and 2022 CIS proposal
%Lumbar Spinal Injection is vital to the diagnosis and treatment of low back pain. According to xxx, About 80\% of adults experience low back pain at some point in their lifetimes, and it is The most common cause of job-related disability and a leading contributor to missed workdays. At least \$50 billion spent on back pain treatment, and another \$100 billion indirect cost annually. Lumbar Spinal Injection is typically performed in the lower back and pelvis area and require targeting of sub-millimeter nerves, small muscular compartments, and thin anatomic spaces. Therefore, the accurate placement of needle is required.

%CT planning, other modalities as feedback ~\cite{bruners2009electromagnetic}

\section{Finite Element Simulator}
\label{sec:fe_simulator}

\begin{figure}[tb]
  \centering
  \includegraphics[width=\columnwidth]{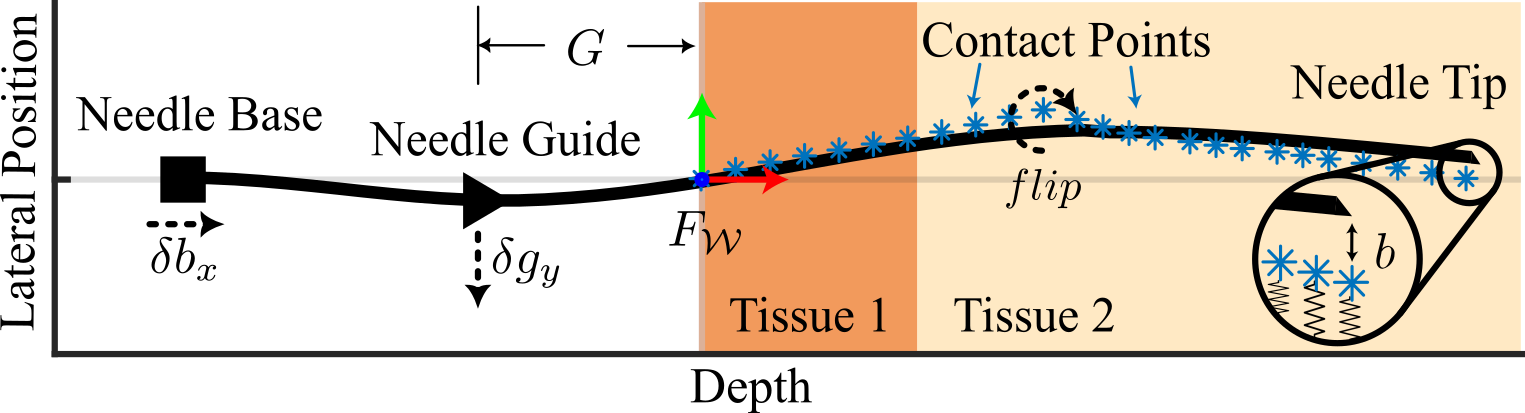}
  \caption{Model schematic generated by the simulator. Three actions, $\delta b_x$, $\delta g_y$, and $flip$ correspond to needle insertion, needle guide translation, and bevel direction change, respectively. Needle guide is separated from skin entry point by distance $G$.}
  \label{fig:simulator}
\end{figure}

Bevel needle manipulation in multi-layered soft tissue environment is simulated in a mechanics-based FE simulator~\cite{Wang2024_2}. The simulator assumes a one-term Ogden hyperelastic model for soft tissue layers. Needle tip geometry is controlled by placing a ``pre-compressed spring'' at the needle tip with a constant offset $b$ (i.e. setting $b=0$ simulates a symmetric-tip needle, see Fig.~\ref{fig:simulator}). This design choice couples needle bevel effect with tissue properties, and allows continuous simulation across different media, for example, from air into skin and muscle.

%, which decouples needle bending and insertion into two separate problems. For bending, the simulator assumes a linear beam model for the needle and a hyperelastic model for biological soft tissues, and allows simulating an arbitrary number of tissue layers with varied biomechanical properties; for insertion, the simulator assumes a constant beam element size so that the length of the needle remains constant, since insertion of a needle should not change the needle length in any way. Needle bevel effect during insertion is achieved by placing a ``pre-compressed spring'' at the needle tip with a constant offset $b$, which can be related to the needle tip geometry (i.e. setting $b=0$ simulates a symmetric-tip needle). This design choice couples needle bevel effect with tissue properties, and allows continuous simulation across different medium (for example, from air into skin and muscle).

The state of the simulation is maintained by a tuple of needle state $X$ and a list of contact points $C$. Evolution of the nonlinear system is represented by a difference equation
\begin{equation}
  \label{eq:difference_equation}
  \left( X_{k+1}, C_{k+1}\right) = f \left( X_k, C_k, U_k \right),
\end{equation}
where $k$ represents the simulation step. 

For an $n-$node discretization of the needle, needle state $X = \left[ x_1, y_1, \theta_1, \cdots, x_n, y_n, \theta_n, bvl \right]^{\top} \in \mb{R}^{3n}\times \mb{Z}_2$ consists of $n$ needle nodal positions and orientations, with a binary variable $bvl = \{-1, 1\}$ indicating bevel direction. The list of contact points, $C$, is dynamically updated at every simulation step based on needle configuration.

Control inputs currently supported by the simulator are $U = \left[ \delta b_x, \delta b_y, \delta b_\theta, \delta g_y, flip \right]^{\top} \in \mb{R}^4\times \mb{Z}_2$, which denote needle (b)ase translations in the $x$ and $y$ directions and rotation $\theta$, needle (g)uide translation in the $y$ direction, as well as a binary signal $flip$ to change bevel direction.

%Control input $U = \left[ \delta b_x, \delta b_y, \delta b_\theta, \delta g_y, flip \right]^{\top} \in \mb{R}^4\times \mb{Z}_2$ consists of continuous needle base manipulations $[\delta b_x, \delta b_y, \delta b_\theta]^{\top}$, movable needle guide translation $\delta g_y$, and a binary signal $flip$ that indicates whether to change the bevel direction. The list of contact points, $C$, is dynamically updated based on the state of the needle tip and carried over to the next simulation iteration. %See Fig.~\ref{fig:simulator} and~\cite{Wang2024_2}.

The complete set of inputs generates redundancy in terms of needle configuration, and can trivialize planning problems. For example, a rigid transformation of the needle can be achieved by moving $\delta b_y$ and $\delta g_y$ in the same direction and by the same amount. Here, we choose a subset of three distinct and representative control inputs that affect needle trajectory in unique ways: $U = \left[ \delta b_x, \delta g_y, flip \right]^{\top} \in \mb{R}^2\times \mb{Z}_2$. Signal $\delta b_x$ moves the entire needle in the direction of insertion, and the action follows the needle base during insertion; $\delta g_y$ introduces needle bending by displacing part of the needle that is in contact with the movable guide, but the point of action is fixed relative to the insertion direction; finally, $flip$ is a binary signal and affects the needle shape only when the needle is in contact with soft tissue and $\delta b_x > 0$. Together, these three control inputs need to act simultaneously to achieve a minimally invasive insertion. 

\section{Cross-entropy Method}
\label{sec:cross_entropy}
The number of states of the simulator in~\cref{eq:difference_equation} changes dynamically depending on the insertion history, and can involve hundreds of state variables depending on mesh resolution and fineness of contact point formation. We show how the cross-entroy (CE) method is well-suited for working with the above system for both trajectory planning and tracking tasks.
%The cross-entropy (CE) method is a well-suited importance sampling method for motion planning algorithms for nonlinear and high-dimensional systems in robotics~\cite{Kobilarov2012, Suh2017, Chen2023}. %The basic idea is to recursively sample from a distribution and update the distribution using a subset of ``elite'' samples with lower costs until the set of samples converges

\begin{figure*}[t]
  \centering
  \includegraphics[width=\textwidth]{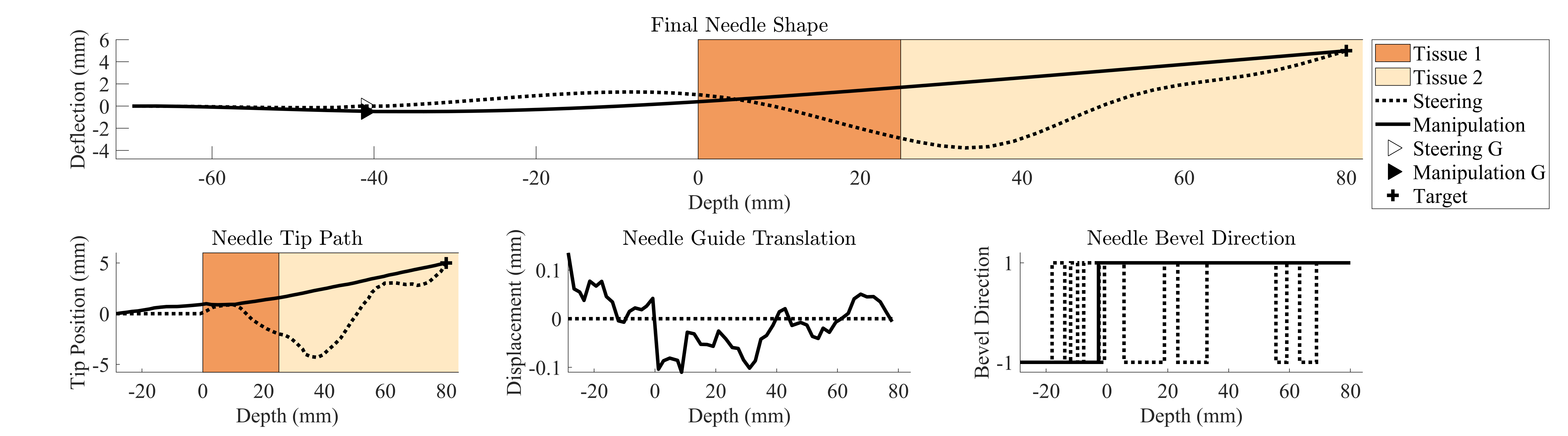}
  \caption{Insertion plans returned by CE using \textit{manipulation} and \textit{steering} strategies. The \textit{manipulation} strategy attempts to find a non-zero insertion angle such that needle bevel can be effectively used to steer towards target, while a pure \textit{steering} strategy relies only on changing of bevel direction to achieve steering.}
  \label{fig:two_plans}
\end{figure*}

Following the development of~\cite{Kobilarov2012}, we consider a needle insertion trajectory as a discretized sequence of $(m + 1)$ states and controls over a fixed time interval $\tau = [0, T]$. From a given initial needle state $X_{0}$, a control sequence $U: \tau\to \mc{U}$ forms a sequence of system states $X: \tau\to \mc{X}$ based on the system difference equation~\ref{eq:difference_equation}. The trajectory is denoted as $\pi: \tau\to \mc{U}\times \mc{X}$, with associated cost $J(\pi)$ defined in general as
\begin{equation}
  \label{eq:general_cost}
  J(\pi) = \sum_{s = 0}^m \phi \left( \pi(s) \right)dt,
\end{equation}
where $\pi(s) = \left( U(s), X(s) \right)$ denotes a tuple of control-state pair at step $s$, $\phi(\cdot)$ a non-negative cost function, and $dt$ a fixed time interval such that $s\cdot dt \in \tau$. In this work, we consider control ``primitives'' $z = \left\{ U_s \right\}$ over $dt$, where each $U_s$, $ 0 \leq s \leq m$, is constant. We parameterize the importance density of $z$ with a Gaussian distribution $p(z, v)$ from the parametric space $\mc{V} = \mb{R}^{n_z}\times \mb{R}^{(n_z^2 + n_z)/2}$ and with element $v = \left( \mu, \Sigma \right) \in \mc{V}$. In other words, independent and identically distributed (i.i.d.) random samples of control primitives $Z_1, \cdots, Z_N$ are drawn from the distribution $p(z, v)$, with mean $\mu$ and covariance $\Sigma$.

The CE approach executes a multi-level search based on the following:
First, initialize counter $j=1$, parameter $v_0 = \left(\mu_0, \Sigma_0 \right)$ and cost threshold $\gamma_0 = \infty$, then
\begin{enumerate}
\item Sample $Z_1, \cdots, Z_N$ from $p(z, v_{j-1})$ and form a set of ``elite'' samples $\mc{E}_j = \left\{ Z_i \big\vert J(Z_i) \leq \gamma_{j-1}\right\}$.
\item Calculate the $\rho$th quantile $\gamma_j = J_{\lceil{\rho N}\rceil}$ for a user-defined parameter $\rho$.
\item Update components of $v_{j+1}$ based on~\cref{eq:update_law_general}.
\end{enumerate}
\begin{equation}
  \label{eq:update_law_general}
  v_j = \argmax_v \frac{1}{N}\sum_{i = 1}^N\mb{I}_{J(Z_i) \leq \gamma_j}\log p(Z_i, v_{j-1}),
\end{equation}
where $\mb{I}_{(\cdot)}$ is the indicator function.
This process is repeated until convergence or when the algorithm reaches the maximum number of execution.

We consider a Gaussian distribution, therefore~\cref{eq:update_law_general} becomes~\cref{eq:update_law_1,eq:update_law_2}.
\begin{align}
  \mu_j &= \frac{1}{\vert\mc{E}_j\vert} \sum_{Z_k \in \mc{E}_j}Z_k, \label{eq:update_law_1}\\
  \Sigma_j &= \frac{1}{\vert\mc{E}_j\vert}\sum_{Z_k \in \mc{E}_j}\left( Z_k - \mu_j \right)\left( Z_k - \mu_j\right)^{\top}. \label{eq:update_law_2}
\end{align}
 
\subsection{Offline Surgical Planning}
\label{sec:trajectory_planning}
For a minimally invasive needle insertion, we desire high targeting accuracy with small tissue compression and few bevel rotations to prevent unnecessary tissue tearing and removal. For a single insertion, the planning cost $J^p$ is the summation of individual step costs $J_s$ and the final cost $J_{f}$, and~\cref{eq:general_cost} takes the form
\begin{equation}
  J^p(\pi) = J_{f}(\pi_f) + \sum_{s=0}^mJ_s(\pi_s).
\end{equation}
The step cost is defined in quadratic form
\begin{equation}
  \label{eq:plan_step_cost}
  J_s(\pi_s) = \left( X_s^{\top}QX_s + U_s^{\top}RU_s \right)\mc{H}(X_s)dt,
\end{equation}
where $Q$ penalizes large needle deflections from an initially straight profile, $R$ penalizes needle manipulations that can potentially be harmful to soft tissues, particularly needle guide translation and bevel direction change. Binary action $flip$ can be optimized by taking the sign of a nonzero numerical value, $\epsilon$, and $flip = \sign({\epsilon})$.

The step cost is activated by a Heaviside function $\mc{H}(X)$ when the needle comes into contact with soft tissue. Final cost is defined at the end of the trajectory, and weighted by constants $\gamma_1, \gamma_2$ between targeting accuracy $\Vert P_{tip} - P_{target}\Vert_2$ and tissue strain energy $\mc{W}$:
\begin{equation}
  \label{eq:plan_final_cost}
  J_{f}(\pi_f)= \gamma_1\Vert P_{tip}(X) - P_{target}\Vert_2 + \gamma_2\mc{W},
\end{equation}
where $P_{tip}$ and $P_{target}$ denote needle tip position and target position, respectively. Tissue strain energy $\mc{W}$ can be calculated using the one-term Ogden hyperelastic model and tissue properties described in~\cite{Wang2023_1}. In short, since the simulator keeps track of a list of active contact points $C$,
\begin{equation}
  \label{eq:ogden}
  \mc{W} = \sum_{c \in C}w_c \frac{2\mu_c}{\alpha_c^2}\left( 2\lambda_{1,3}^{\alpha_c} + \lambda_2^{\alpha_c} - 3 \right),
\end{equation}
for each contact point $c$. Tissue model parameters $\mu, \alpha$ are defined on a per-layer basis based on the location of contact points. Stretch ratios $\lambda_i$ are calculated at each contact point with respect to the corresponding needle deflection. Unconfined uniaxial compression is assumed and $\lambda_2$ is assigned as the principal stretch ratio. Constants $w_c$ are optional weighting factors for different tissue layers.

Planning phase returns a minimally invasive trajectory $\bar{\pi}$, that consists of an optimal control sequence $\bar{U}$ and associated needle states $\bar{X}$. These nominal values are used to warm start the online trajectory tracking algorithm derived from the same CE framework. 

A kinematic-like \textit{steering} plan can be trivially obtained by lowering the needle bending stiffness in the simulator and zeroing the covariance entries that correspond to inputs that actively change the needle shape. A comparison of optimal plans of the two strategies is shown in Fig.~\ref{fig:two_plans}. Note that since bevel effect is coupled with tissue stiffness, needle \textit{manipulation} can affect needle shape and trajectory before the needle touches soft tissue. In this work, we focus on needle \textit{manipulation}, but we show in Sec.~\ref{sec:kinematic_models} how to bridge kinematic models for trajectory tracking.

\subsection{Online Trajectory Tracking}
\label{sec:trajectory_tracking}
To track the nominal trajectory subject to uncertainties in initial conditions and model parameters, we extend the CE optimization framework to make a model predictive controller (MPC) for continuous input signal generation. We further couple it with an ``ad hoc'' bang-bang controller to switch bevel tip directions.

\subsubsection{Model Predictive Controller}
\label{sec:mpc}
The CE optimization framework can be extended to build an MPC for online trajectory tracking. The underlying idea is to execute a single control input $U$ sampled from $\bar{U}$ that minimizes tracking error within a short window in a receding-horizon fashion.

We consider a moving window $\tau^{\prime} \subset \tau$ with the same discrete time step $dt$. To warm start the algorithm, we set the new sample mean $\mu^{\prime} = \bar{U}^{\prime} \subset \bar{U}$  within the time window. The covariance matrix $\Sigma^{\prime}$ is re-initiated depending on the noise of the environment and discrepancy of the needle initial conditions, and the algorithm introduced in Sec.~\ref{sec:cross_entropy} is applied to get the next control input.

The nominal trajectory $\bar{X}$ contains not only the needle tip position $P_{path}$, but also the tip orientation $\theta_{path}$ returned by the simulator. For off-the-shelf needles, tip orientation plays an active role in terms of long-term needle trajectory, and is important to track if possible~\cite{Wang2024_3}. Therefore, considering the needle tip state $g(X) = \left( x_n, y_n, \theta_n \right)\in SE(2)$, trajectory tracking cost $J^t$ is the cumulative deviation from the nominal path $\bar{g}(\bar{X})$ over the short horizon $\tau^{\prime}$ according to
\begin{equation}
  \label{eq:exec_cost}
  J^t(\pi, \bar{\pi}) = \sum_{s^{\prime} = 0}^{m^{\prime}}d_{s^{\prime}} \left( g, \bar{g} \right)_Wdt,
\end{equation}
where $s^{\prime}dt \in \tau^{\prime}$, and $d \left( \cdot, \cdot \right)_W : SE(2)\times SE(2) \to \mb{R}$ assigns a cost to the deviation based on weight $W$.

Contrary to planning, the MPC is only in charge of continuous input signals. Binary input signal, $flip$, is generated by a bang-bang controller derived from a kinematic model.

\subsubsection{Kinematic-based Bang-Bang Controller}
\label{sec:kinematic_models}
The necessity of the bang-bang controller is as follows. First, given the current set of $U$, needle tip position and orientation cannot be independently controlled, therefore even if the MPC can correct the initial tip position deviation while the needle is in air, the needle will enter soft tissue at a different angle from the plan, and vice versa. Second, in an environment where the relative stiffness between needle and tissue is high, bevel effect is diminished until the inserted depth becomes large and needle motion appears nonholonomic. These two issues present a major challenge to the MPC since inherently a short trajectory window is considered, which can put the needle in an irrecoverable state.

The bang-bang controller assumes a kinematic model using the current needle tip configuration $g(X)$ based on the following discrete kinematic equation
\begin{equation}
  \label{eq:discrete_kinematic}
  \begin{bmatrix} x_{n, k + 1} \\ y_{n, k + 1} \\ \theta_{n, k + 1} \end{bmatrix} =
  \begin{bmatrix} x_{n, k} \\ y_{n, k} \\ \theta_{n, k}  \end{bmatrix} +
  \begin{bmatrix} \cos(\theta_{n, k}) \\ \sin(\theta_{n, k}) \\ sgn\cdot\kappa \end{bmatrix}u,
\end{equation}
where $sgn \in [bvl, -bvl]$ takes either the current bevel direction or its opposite, $\kappa$ is a small positive value that will generate a curved path with direction determined by $sgn$. The continuous insertion signal $\delta b_x$ obtained by the MPC is divided into smaller steps, which are $u$ in the above equation.

Since $sgn$ can only take two values, two needle tip trajectories will be generated. Decision $flip$ is then based on the trajectory that brings the needle tip closer to the final target position. In other words, if we define the tip position at the end of kinematic trajectory generated by choosing bevel direction $sgn$ as $P_{tip}^{sgn}$, then
\begin{align}
  \label{eq:bang_bang}
  bvl^{*} &= \argmin_{sgn} \Vert P_{tip}^{sgn}(X) - P_{target} \Vert_2, \\
  flip &=
  \begin{cases}
     -1, & \text{if } bvl^{*} = bvl \\
     1, & \text{otherwise.}
  \end{cases}
\end{align}

The reason for using $P_{target}$ instead of $P_{path}$ in~\cref{eq:bang_bang} is to introduce long-horizon prediction so that the bevel tip always points in the direction of target, instead of chasing the nominal path by introducing frequent bevel direction changes.

\section{Controller Feedback}
\label{sec:controller_feedback}
\begin{figure}[tb]
  \centering
  \includegraphics[width=\columnwidth]{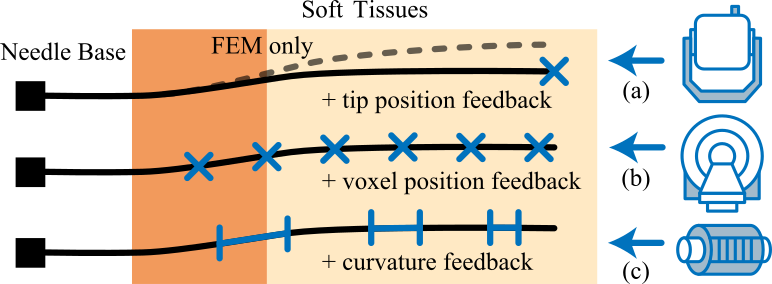}
  \caption{Comparison between different feedback modalities: (a) EM tracking, (b) medical imaging, and (c) fiber Bragg grating with the proposed FE-based framework.}
  \label{fig:feedback_modalities}
\end{figure}

Common feedback modalities that fit within the proposed control framework include electromagnetic (EM) tracking, medical imaging, as well as FBG. An EM sensor can be attached to the needle to provide a single position and orientation feedback in high frequency, while medical imaging can provide approximate positional information along the length of the needle, subject to image resolution, speed of acquisition, and accuracy of image segmentation. FBG sensors can be embedded in the needle to provide partial shape information in real time~\cite{Wang2024_1}. A pictorial comparison between different feedback modalities compatible with the presented framework is shown in Fig.~\ref{fig:feedback_modalities}.

\subsection{Position Feedback with EM Tracking}
\label{sec:position_feedback}
In this work, we use an EM sensor coil embedded at the tip of an 21 gauge needle with 0.819mm outer diameter (Aurora, NDI Digital, Canada) to provide position feedback, which is treated as an essential boundary condition in the FE solver. Although it is possible to incorporate sensor orientation information, considering the generalizability of proposed approach to medical imaging, orientation would be challenging to extract precisely. Therefore, only tip position is used as feedback.

\subsection{EM Noise and Accuracy Evaluation}
\label{sec:em_evaluation} 

The choice of using EM tracking is motivated by its speed and accuracy within a ``good'' measurement window; however, EM sensors are vulnerable to field distortions introduced by components of the needle driver and devices commonly found in the operating room (OR). While the operating environment recommended by the manufacturer should be clear of any ferromagnetic object within 1m of the EM field generator~\cite{NDI_2013}, such requirement is unrealistic in any OR. 

In order to define an operating window for position tracking and to incorporate the tracking device into our setup, an evaluation of the accuracy of the 5-degree-of-freedom (DOF) needle tip tracker and 6DOF reference trackers (Aurora, NDI Digital, Canada) is performed.

We attach both the 5DOF tip tracker and a 6DOF reference tracker onto a motorized XYZ linear stage assembly. By repeatedly adjusting the positions of the two sensors with the linear stage and calculating the measurement error, we find that a minimum distance of 70cm between the field generator and the linear stage is required to achieve an overall root mean square error (RMSE) of 0.17mm and 0.25mm for the two sensors, respectively, which are lower than the manufacturer-reported RMSE of 0.70mm and 0.48mm~\cite{NDI_2013}. Since no other metallic and power sources are present in the setup, we call this the \textit{ideal} environment.

\begin{figure}[t]
  \centering
  \includegraphics[width=\columnwidth]{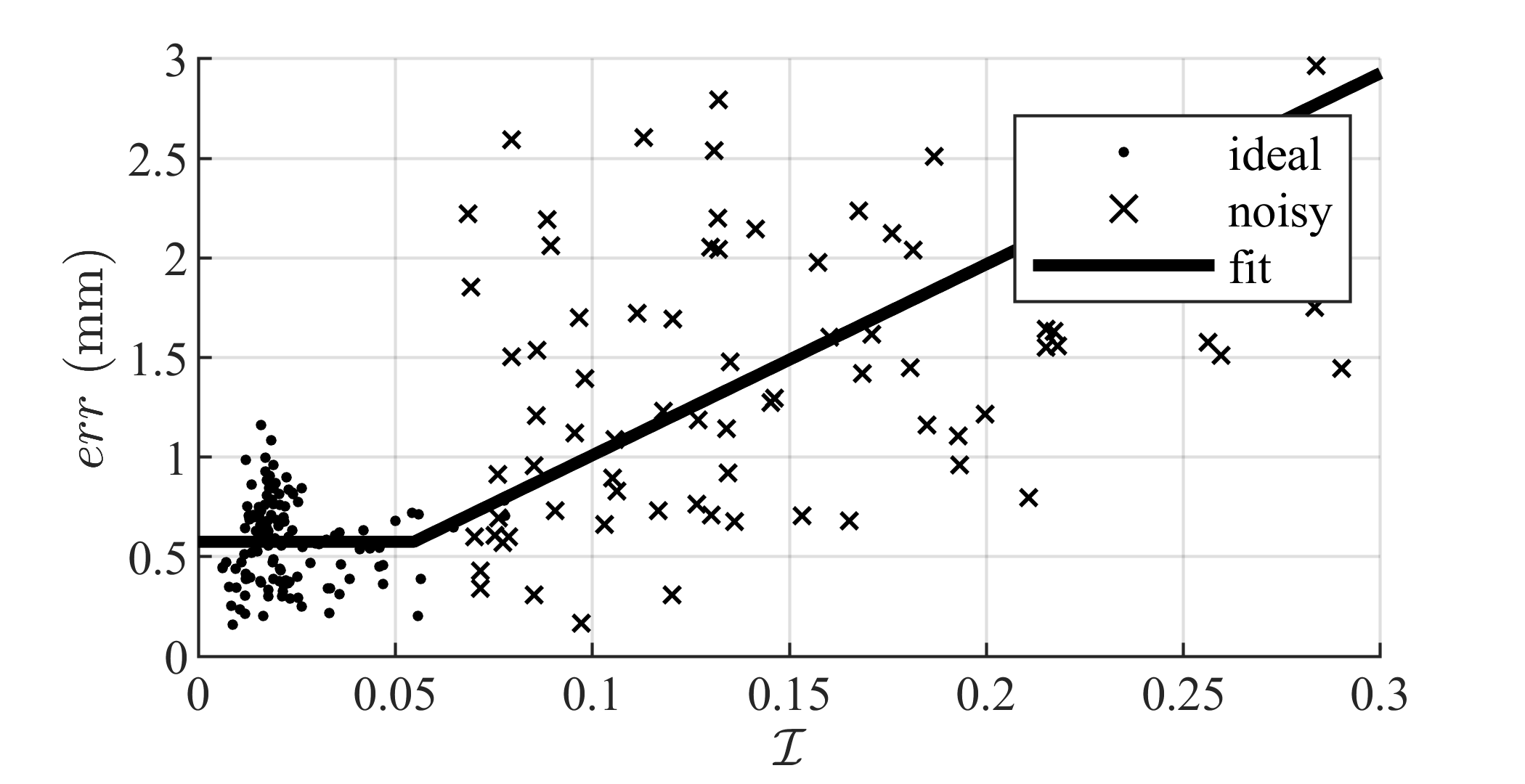}
  \caption{Indicator value correlates measurement error in a piecewise linear fashion. Indicator values beyond 0.3 are not shown to better visualize data in the \textit{ideal} environment.}
  \label{fig:indicator_value}
\end{figure}

Each 6DOF reference tracker consists of two 5DOF sensors and returns an indicator $\mc{I}$ from 0 to 9.9 to indicate field distortion. To find the correlation between measurement error and $\mc{I}$,  we use the linear stages to move the sensors on an $5 \times 5 \times 5$ square grid within a 1000cm$^3$ volume, with 100 position measurements collected at each grid point. The data collection process is repeated for 5 times, therefore a total of $M = 500$ measurements are collected at each of $N = 125$ grid points.
% note: should mention this for repeatibility
Furthermore, we deliberately introduce additional field distortion by placing a metal plate underneath the setup. We call this the \textit{noisy} environment.

\begin{table}[b]
\centering
\caption{Sensor Measurement Errors}
\label{tab:sensor_errors}
\begin{tabular}{c|p{0.4cm}p{0.4cm}p{0.4cm}p{0.6cm}|p{0.4cm}p{0.4cm}p{0.4cm}p{0.6cm}}
\toprule
%\hline
\multirow{2}{0.65cm}{Error (mm)} & \multicolumn{4}{c|}{\textbf{5DOF Tip}} & \multicolumn{4}{c}{\textbf{6DOF Ref.}} \\
         % \midrule
    & $rmse$       & 95\%\textsuperscript{a)}  & max  & $jitter$      & $rmse$           & 95\%\textsuperscript{a)}  & max  & $jitter$\\
\midrule
Official & 0.70       & 1.40 & -        &   -     & 0.48           & 0.88 & -        &-         \\
ideal    & 0.65       & 1.10 & 1.40     & 0.07      & 0.61           & 0.91 & 1.16      &0.08 \\
noisy    & 1.53       & 3.20 & 4.36     &  0.04       & 2.70           & 5.27 & 8.16     & 0.07\\
\bottomrule
\end{tabular}
\par\smallskip
 \parbox{\columnwidth}{\raggedright\small
    a) 95\% confidence interval}
\end{table}

For each EM measurement, $\tilde{P}_{i,j}$ represents position at the $i$th grid point out of $N$ grid points, and the $j$th measurement out of $M$ total samples. The mean position of the $M$ samples at the $i$th grid point can be calculated as 
\begin{equation}
  \label{eq:mean_pos}
  \bar{P}_{i} = \frac{1}{M}\sum_{j=1}^M \tilde{P}_{i,j}.
\end{equation}
Since we treat the grid points formed by the linear stage as true positions $P_i$, measurement error is defined as the Euclidean distance between the measured mean position and the ground truth position
\begin{equation}
\label{eq:measurement_error}
err_i =  \Vert \bar{P}_{i} - P_{i}\Vert _2,
\end{equation}
and the RMSE of all $N$ grid points in a given environment can be obtained by
\begin{equation}
  \label{eq:em_rmse}
  rmse = \sqrt{\frac{\sum_{i=1}^N err_i^2}{N}}.
\end{equation}
Similarly, since multiple samples are taken at each grid point, sensor jitter in a given environment can be evaluated with 
\begin{equation}
  \label{eq:em_std}
  jitter = \sqrt{\frac{\sum_{i=1}^N \sum_{j=1}^M \Vert \bar{P}_{i} - \tilde{P}_{i,j} \Vert_2^2}{NM}}.
\end{equation}

As shown in Table~\ref{tab:sensor_errors}, in the ideal environment, the RMSE is consistent with the official product specifications~\cite{NDI_2013}. When the metal plate is introduced and environment becomes noisy, the respective maximum errors of the two sensors rise to 4.36mm and 8.16mm, but jitter does not have a significant difference in the noisy environment. The reason for the accuracy of the tip being higher than that of the reference can be attributed to the needle tip being closer to the field generator in our setup~\cite{Nakamoto2000}.

% \begin{table}[]
% \caption{error of needle tip}
% \label{tab:tip_err}
% \begin{tabular}{lllll}
%          & RMSE & P95  & max  & jitter \\
% ideal    & 0.65 & 1.10 & 1.40 & 0.07   \\
% noisy    & 1.53 & 3.20 & 4.36 & 0.04   \\
% official & 0.70 & 1.40 & -    & -     
% \end{tabular}
% \end{table}

% \begin{table}[]
% \caption{error of 6DOF reference}
% \label{tab:ref_err}
% \begin{tabular}{lllll}
%          & RMSE & P95  & max  & jitter \\
% ideal    & 0.61 & 0.91 & 1.16 & 0.08   \\
% noisy    & 2.70 & 5.27 & 8.16 & 0.07   \\
% official & 0.48 & 0.88 & -    & -     
% \end{tabular}
% \end{table}

% In an ideal environment, $err_{RMSE}$ is 0.65 mm for the needle tip, and 0.61 mm for the 6DOF reference; with the maximum error being 1.40 mm and 1.16 mm respectively. The results are consistent with the product technical specifications, whose RMSE is 0.70 mm for the tip and 0.48 mm for the reference~\cite{NDI_2013}. While when noise is introduced, the respective maximum errors will rise to 4.36 mm and 8.16 mm. $err_{jitter}$ does not have a significant difference in two environments, 
% %and does not have a strong correlation with the error value, 
% being 0.05 mm for the tip, and 0.08 mm for the reference. The reason for the accuracy of the tip being higher than that of the reference, contrary to the official specifications, is due to the needle tip being closer to the field generator in our setup~\cite{Nakamoto2000}. 

As shown in Fig.~\ref{fig:indicator_value}, a piecewise linear fit suggests that in the ideal environment, $err = 0.575\pm 0.211$mm; in the noisy environment, $err = 9.591\mc{I} + 0.051$mm. We take the indicator value $\mc{I} = 0.0547$, which is the 95 percentile of $\mc{I}$ measured in the ideal environment, as a good representation of a rather noise-less environment. This choice is further corroborated by Table~\ref{tab:indicator_value}, which shows that in the noisy environment, $\min \mc{I} > 0.0547$. 
% note: the threshold should be kept low, 0.06 is too close to min of noisy values; also in real experiments Is are always below 0.05, so I think using 95% percentile is more reasonable
This threshold value is used in our setup, where three reference markers are placed around the tissue phantom in order to estimate noise magnitude around the needle workspace, and keep all three indicator values below the threshold. 

%\textcolor{red}{The 6DOF reference will return a unitless indicator value $i$ whose range is 0 - 9.9, to indicate its EM distortion. Since an Aurora 6DOF sensor consists of two 5DOF sensors, such a value is calculated by comparing the design distance and measured distance between the two components. However, NDI does not provide a relationship between this value and the physical distortion distance. In the above experiment, we found that 
%in the ideal environment, the indicator value is below 0.06, while when noise is introduced, it will surpass 0.06 and can be up to 0.5, and 
%the two are directly correlated, as seen in Fig.~\ref{fig:indicator_value}. 
%We take the indicator value of $\mc{I} = 0.0547$, the 95\% percentile of $i$ measured in the ideal environment, as a threshold $thr$ to indicate an ideal environment, since such a value as a good representation of the possible $\mc{I}$ when no noise is present, also, the minimum $\mc{I}$ in an noisy environment is measured to be 0.0684. Therefore, $thr$ can be used to separate a noiseless environment, where distortion can be formulated as $d = 0.575 \pm 0.211 mm$, from a noisy environment, where $d = 9.591 \mc{I} + 0.051 mm$.}
%Therefore, when the indicator value of a tracker is kept under a threshold of 0.06, 
%then, the system will have an accuracy and jittering sufficient for our task. 
% indicator value:
% ideal:    0.0239 mean,    0.0138 stddev,  0.0779 max,  0.0061 min
% aluminum: 0.1944 mean,    0.1049 stddev,  0.4784 max, 0.0684 min

\begin{table}[t]
\centering
\caption{Indicator Value, $\mc{I}$}
\label{tab:indicator_value}
\begin{tabular}{c|c|c|c|c}
\toprule
$\mc{I}$     & mean   & sd & max    & min    \\
\midrule
ideal & 0.0239 & 0.0138 & 0.0779 & 0.0061 \\
noisy & 0.1944 & 0.1049 & 0.4784 & 0.0684 \\    
\bottomrule
\end{tabular}
\end{table}

\section{Phantom Experiment with EM Feedback}
\label{sec:phantom_experiment}

\begin{figure}[b]
  \centering
  \includegraphics[width=\columnwidth]{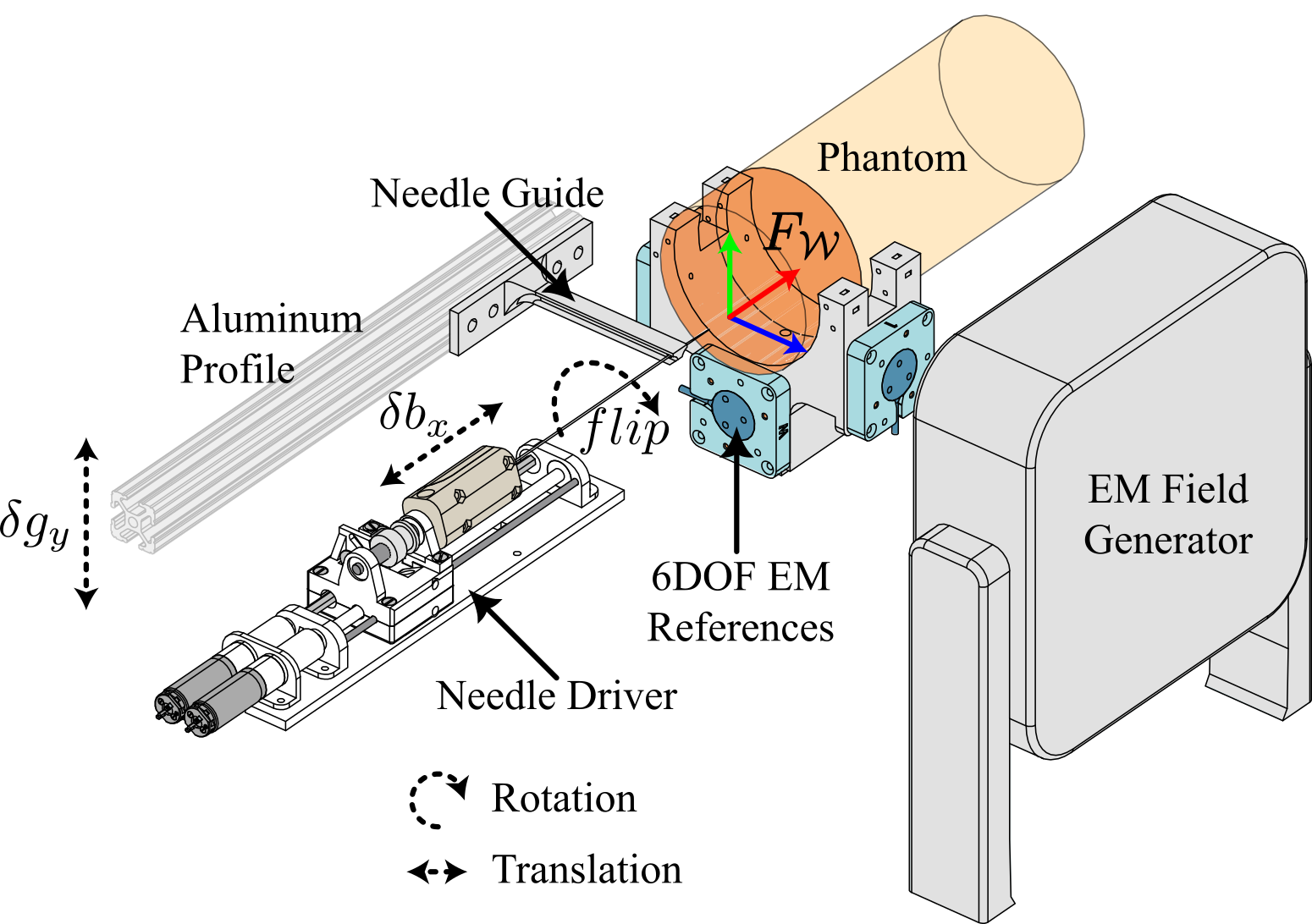}
  \caption{Tissue phantom experiment setup. Needle driver generates $\delta b_x$ and $flip$, and linear stage (not shown) generates needle guide $\delta g_y$ attached to the aluminum profile. Three EM reference markers are used to monitor field distortion.}
  \label{fig:setup}
\end{figure}

\begin{figure*}[t]
  \centering
  \includegraphics[width=\textwidth]{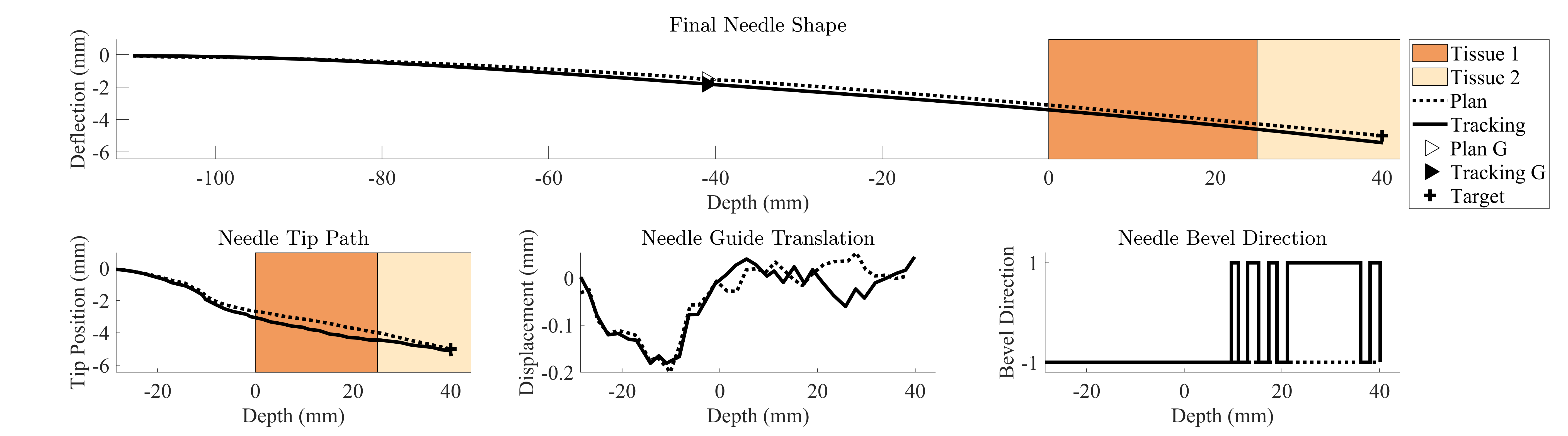}
  \caption{Needle insertion planning and tracking for target located 40mm deep with 5mm offset from needle initial configuration.}
  \label{fig:example_em_insertion}
\end{figure*}

Phantom experiments are performed with hardware setup shown in Fig.~\ref{fig:setup}. The 2DOF needle driver provides translation distance up to 110mm, with rotation limited to $0^{\circ}$ and $180^{\circ}$ to replicate the simulator setup. The movable needle guide is actuated by the same linear stage used in Sec.~\ref{sec:em_evaluation}, and is attached to an aluminum profile in order to reduce electromagnetic interference. All motors are controlled independently by a motion controller (DMC-4143, Galil Motion Control, USA), and cross-platform communication is established using ROS2~\cite{Macenski2022}. When the hardware system is powered on and placed in an ideal environment, $\mc{I} = 0.04$, proving EM compatibility of our system. 

\begin{figure}[b]
  \centering
  \includegraphics[width=\columnwidth]{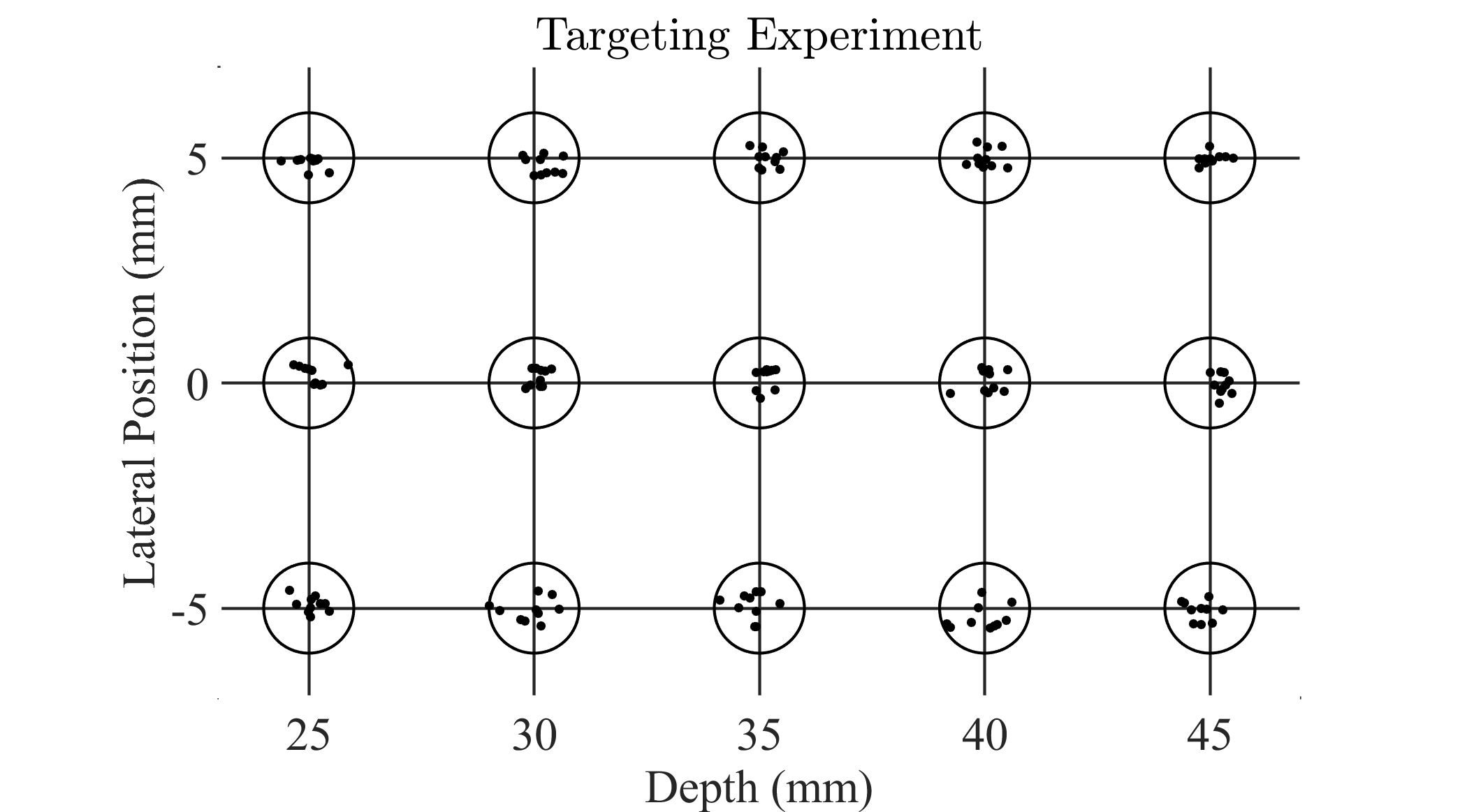}
  \caption{Targeting experiment with guide-to-skin distance $G=41$mm. Targets are shown as the center of circles, which have radius of 1mm. Black dots show the needle tip position reported by EM tracker.}
  \label{fig:em_error}
\end{figure}

Tissue phantom used in the experiment has two layers of different stiffness values. The superficial and deep layers are assigned Ogden model parameter $\mu_1 = 1.82$kPa and $\mu_2 = 3.63$kPa, respectively, with $\alpha = 8.74$ for both layers~\cite{Wang2024_1}. Needle Young's modulus is set to 200GPa for stainless steel. The needle guide is positioned at $G=41$mm with initial tip-to-skin distance of $28.5$mm. Target depths are chosen to be from $25$mm to $45$mm in $5$mm increments to simulate a lumbar injection scenario, where the average insertion depths is $36.7 \pm 9.9$mm~\cite{Fritz2012}. With the 21 gauge bevel tip needle initially aligned with $0$mm lateral axis, we also target positions with $\pm 5$mm lateral variations to simulate initial registration errors and misalignment. Fifteen targets in total are selected and ten insertions are performed on each target.

\section{Results and Discussion}
\label{sec:results_and_discussion}

An example insertion for the proposed planning and control framework is shown in Fig.~\ref{fig:example_em_insertion}. %The controller is able to manipulate the needle such that it follows the planned trajectory well and reaches the final target with high accuracy. 
Notably, for targets with large offset from the needle initial configuration, the proposed framework is able to take advantage of the gap between needle tip and entry point to perform the most correction in air, and the controller is able to generate needle guide motions that closely resembles the plan.

Once the needle touches the tissue, the CE-driven MPC and kinematic-driven bang-bang controller is able to work together, adjusting the guide position and bevel direction as needed. There are, however, more bevel direction changes in close proximity with each other, which can be undesirable depending on the specific clinical scenario. This is in part due to the needle not rotating axially in place such that changing the bevel direction also changes the tip position. Reducing the curvature value $\kappa$ or implementing an error threshold in~\cref{eq:bang_bang} can mitigate this issue.

Through experimentation, we decide that needle orientation upon entering the tissue is more important than position, particularly in determining long-term tracking performance. Since the set of control inputs considered in the present case (see Fig.~\ref{fig:simulator}) cannot independently control tip position and orientation, higher weight is given to needle tip orientation, which explains the non-zero tracking error at the tip near the needle entry point. The overall performance is shown in in Fig.~\ref{fig:em_error}. Target positions are centers of the circles, which have radius of 1mm. Final needle tip positions are represented as dots. The average targeting error is $0.16\pm0.29$mm based on the final position returned by the tip tracker.

\section{Conclusion and Future Work}
\label{sec:conclusion_and_future_work}
We demonstrate an effective integration of the CE method with a real-time FE simulator for minimally invasive needle manipulation planning and tracking. Our approach leverages a CE-driven MPC in conjunction with a kinematic-driven bang-bang controller to achieve precise needle insertions. Tissue phantom experiment reveals targeting accuracy is within 1mm of various targets. Future research will focus on extensive experiments with real soft tissues and needles of different gauges, as well as impacts of variations in guide-to-skin distance and fusion with medical images.

\bibliography{bibliography}
\bibliographystyle{ieeetr}

\end{document}